\documentclass{article}

\usepackage{microtype}
\usepackage{graphicx}
\usepackage{subcaption}
\usepackage{booktabs} 

\usepackage{hyperref}

\usepackage[accepted]{icml2026}

\usepackage{amsmath}
\usepackage{amssymb}
\usepackage{mathtools}
\usepackage{amsthm}

\usepackage[capitalize,noabbrev]{cleveref}

\theoremstyle{plain}

\theoremstyle{definition}

\theoremstyle{remark}

\usepackage[textsize=tiny]{todonotes}

\usepackage{makecell}

\icmltitlerunning{Decoding Insect Song: A Multitask Semisupervised Orthoptera Bioacoustic Classifier}

\begin{document}

\twocolumn[
  \icmltitle{Decoding Insect Song: A Multitask Semisupervised Orthoptera Bioacoustic Classifier}



  \icmlsetsymbol{equal}{*}

  \begin{icmlauthorlist}
    \icmlauthor{Olga Isupova}{equal,geog}
    \icmlauthor{Danil Kuzin}{equal,geog}
    \icmlauthor{Ella Browning}{geog}
    \icmlauthor{Tom Mills}{geog}
    \icmlauthor{Steven Reece}{geog}
  \end{icmlauthorlist}

  \icmlaffiliation{geog}{Leverhulme Centre for Nature Recovery, School of Geography and the Environment, University of Oxford, Oxford, UK}
  
  \icmlcorrespondingauthor{Olga Isupova}{olga.isupova@ouce.ox.ac.uk}
 
  \icmlkeywords{multitask learning, semisupervised learning, audio processing, orthoptera classification}

  \vskip 0.3in
]



\printAffiliationsAndNotice{\icmlEqualContribution}

\begin{abstract}
 Passive acoustic monitoring holds great promise for ecological inference, yet existing automated tools are typically narrowly trained and non-transferable. We address these limitations with PULSE, a semi-supervised, multi-task framework for Orthoptera bioacoustics, combining weakly-supervised species classification, self-supervised learning on unlabelled field audio, and knowledge distillation from a general-purpose bioacoustic model. Our domain-adapted specialist model outperforms a state-of-the-art general model across all metrics (macro F1: 0.21 vs. 0.07; AUC: 0.74 vs. 0.45; AP: 0.32 vs. 0.19), with active learning further raising F1 to 0.34 and AUC to 0.84. Beyond classification, the learned embeddings encode ecologically meaningful structure, exposed through an interactive visualisation tool for ecological discovery.
\end{abstract}

\section{Introduction}

 Passive acoustic monitoring (PAM) is increasingly used to collect ecoacoustic data to assess habitat health and biodiversity responses to environmental change~\cite{gibb2019emerging, ross2023passive}. By extracting taxa-specific signals, spatial and temporal patterns in species presence, activity, and density can be identified, informing land management and conservation strategies. Automating this using machine learning (ML) has enabled large-scale PAM for birds \cite{kahl2021birdnet, hamer2023birb} and bats \cite{aodha2022towards}. However, despite the ecological importance of insects, there are no open-source ML tools for acoustically identifiable insects such as Orthoptera (crickets and grasshoppers), which use songs and calls for various ecological functions \cite{riede2025acoustic}. A key barrier is the lack of large open-access sound libraries to train ML algorithms, though new datasets on platforms such as Xeno-canto~\citep{vellinga2024xeno} and the Global Biodiversity Information Facility are beginning to address this \cite{funosas2026finely}. Combining these data with ML approaches that reduce the need for large training sets, such as agile modelling and self-supervised learning, presents meaningful opportunities to address PAM taxonomic biases.
 
Extraction of Orthoptera signals from ecoacoustic data is more challenging than for birds or bats: many species call near-continuously whilst active, and overlapping calls are common \cite{funosas2026finely}. Furthermore, reference calls on bioacoustic platforms are often unrepresentative of PAM recordings, with higher audio quality and limited noise, causing ML tools to perform poorly on field recordings. Combining ``clean'' sound library data with ``messy'' field recordings can improve model performance. To address these challenges, we present PULSE, which leverages a large collection of unlabelled UK field recordings --- released alongside this work --- to bridge the gap between curated sound libraries and real-world deployment. 

 \section{Related Work}
Recent approaches for Orthoptera PAM are based on deep learning. OrthopterOSS~\citep{bennett2025recent} uses the AMResNet architecture~\citep{xiao2022amresnet} but, like earlier models~\citep{faiss2023adaptive,newson2017potential}, relies solely on supervised classification. We argue that supplementing training with unsupervised tasks is essential to overcome the scarcity of labelled Orthoptera data.


Large models that were trained on soundscape datasets, such as BirdNET~\citep{kahl2021birdnet} and Perch~\citep{hamer2023birb}, have demonstrated good performance on birds. Embeddings extracted from these models capture general-purpose audio representations from soundscapes and are useful for training downstream tasks. We distilled generalised knowledge from BirdNET by matching its embeddings.

Models trained on large global datasets often have lower performance on local datasets due to domain shift. Using multi-task training~\cite{caruana1997multitask}, and combining supervision for known taxonomic classification, distillation for general ecoacoustic knowledge, and self-supervised learning (SSL) for local acoustic profile, we address the data scarcity and domain shift problems. 

\section{Methodology}
We propose Passive acoUstic Latent-Space Encoder (PULSE): a multi-task framework that includes unsupervised and supervised training. 

\subsection{Data}
We used publicly available labelled data for Orthoptera and supplemented it with a large, unlabelled dataset of UK field recordings. A small subset of field recordings included labels provided by experts through active learning.

For our labelled data we used the recently compiled ECOSoundSet, which collates Orthopteran and other ecoacoustic data~\citep{funosas2026finely}, and online datasets such as Xeno-canto~\citep{vellinga2024xeno} and iNaturalist, collected by the script provided by the authors of~\citet{funosas2026finely}. Refer to Appendix~\ref{app:data} for details. We selected a subset of 19 Orthoptera species from these sources that are endemic to, or present in the UK (Table~\ref{tab:species_list}).

We supplemented the labelled data with over 130 GB collection of unlabeled field recordings collected across 10 sites in Oxfordshire, UK. We have released this data together with labels collected with active learning (Section \ref{sec:active-learning}). It is available at \hyperref[https://doi.org/10.5281/zenodo.20625749]{https://doi.org/10.5281/zenodo.20625749}. 

\subsection{Backbone}
First, an audio is transformed into a mel spectrogram. To process input spectrograms into embeddings $e$ we use a VGGish~\citep{hershey2017cnn} backbone. The architecture was modified to match our spectrogram shape for input and BirdNET embedding size for output since we used BirdNET as ecologial prior.


\subsection{Multi-task Framework}
\begin{figure}[tb]
  \centering
  \includegraphics[width=\columnwidth]{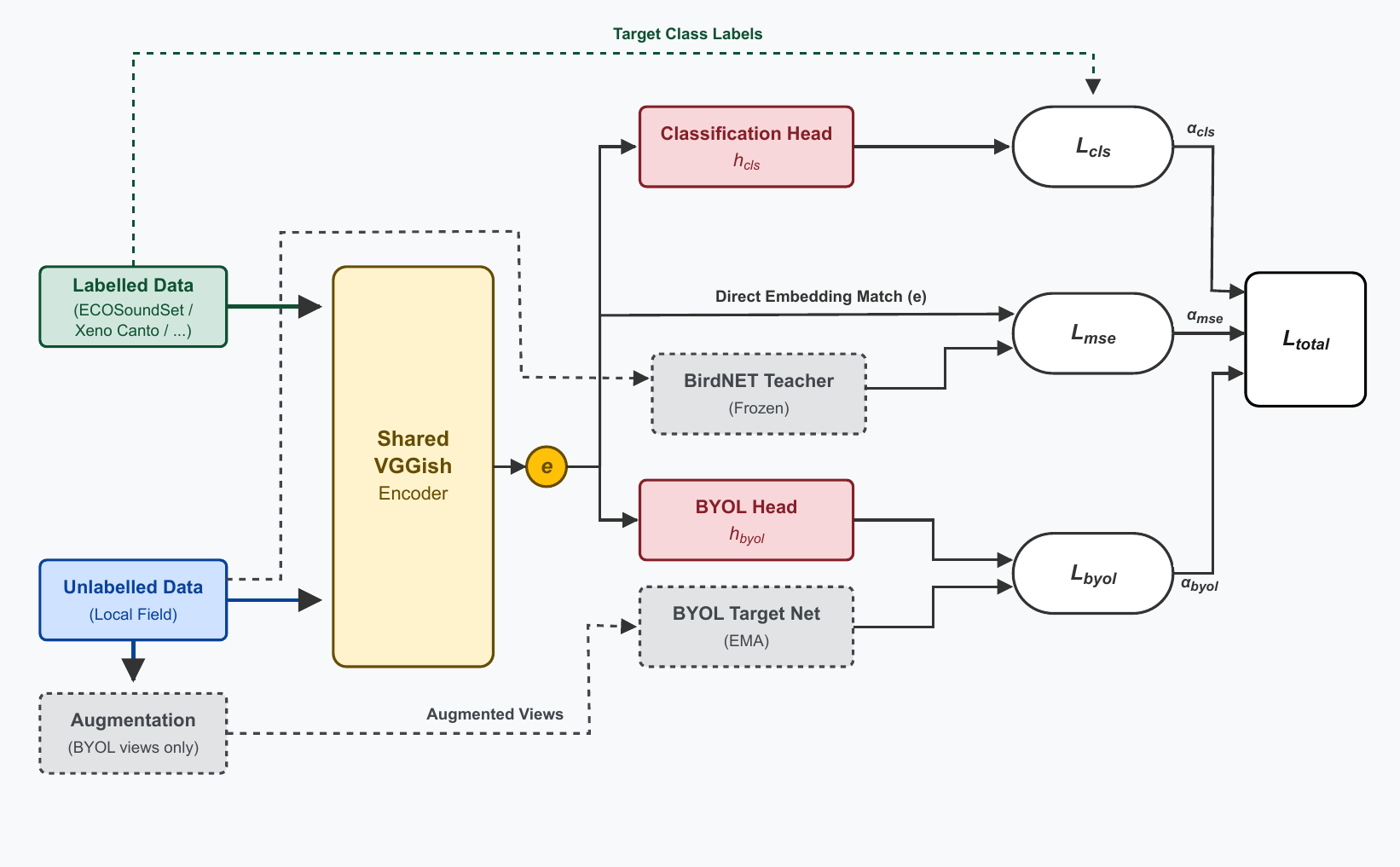}
  \caption{PULSE framework architecture with three tasks.}
  \label{im:architecture}
\end{figure}

Backbone embeddings serve as an input to our multi-task framework, that combines three optimisation objectives. \figurename~\ref{im:architecture} shows the framework architecture.

\textbf{Supervised classification.}
Using available labelled data, we optimised the classification head $h_{\text{cls}}$. Since recordings can contain several species vocalising at the same time, we have a multilabel problem, therefore, $h_{\text{cls}}$ has the binary cross-entropy loss $\mathcal{L}_{\text{cls}}$. 

\textbf{Ecological prior.}
The pretrained BirdNET model provides useful representations for ecoacoustics. We encouraged our embeddings $e$ to match BirdNET embeddings on unlabelled data. This was achieved by optimising the mean squared $L_2$ loss $\mathcal{L}_{\text{mse}}$ between our embeddings and the frozen BirdNET output. Since we matched BirdNET, we also work with 3-sec input clips.

\textbf{Self-supervision.}
To adopt to our domain soundscapes, $h_{\text{byol}}$ head of the framework used the unlabelled field data and applied the Bootstrap Your Own Latent~\citep{grill2020bootstrap} method.
We augmented spectrograms with time and frequency masking to create two views of the same data. Using the online and target networks, updated with exponential moving average, we optimised the $L_2$ loss $\mathcal{L}_{\text{byol}}$ between the normalised predictions of the online network and the projections of the target network.

\subsection{Joint Optimisation}
During training, we created batches sampled for each of three tasks simultaneously. We then optimised the weighted sum of the three loss functions: 
$
\mathcal{L}_{\text{total}} = \alpha_{\text{cls}}\mathcal{L}_{\text{cls}} + \alpha_{\text{mse}}\mathcal{L}_{\text{mse}} + \alpha_{\text{byol}} \mathcal{L}_{\text{byol}}
$,
where we set $\alpha_{\text{cls}}=0.9$, $\alpha_{\text{mse}}=0.1$, $\alpha_{\text{byol}}=0.1$. These weights accommodate significant imbalance between labelled and unlabelled data.

\subsection{Active Learning}
\label{sec:active-learning}

We used PULSE as a base model for labelling a part of field data with active learning using the DIRECT method~\cite{nuggehalli2025improved} designed for imbalanced data. We followed the standard cycle: train PULSE, predict on unlabelled data, select the most informative batch with DIRECT, manually label it, and repeat. Labelling was conducted using Whombat~\cite{balvanera2025whombat}\footnote{Can be found at \href{https://github.com/mbsantiago/whombat/}{https://github.com/mbsantiago/whombat/}}.



\section{Experiments}

\begin{table}[t]
  \caption{Benchmark models and data they use for training.}
  \label{tab:model-data}
  \begin{center}
    \begin{small}
      \begin{tabular}{m{2cm} m{1cm} m{1cm} m{1cm} m{1cm}}
        \toprule
        Model & PULSE species library & Perch species library & Unla\-belled field & Label\-led field \\
        \midrule
        PULSE & \checkmark &  --- & \checkmark & --- \\
        active PULSE & \checkmark &  --- & \checkmark & \checkmark \\
        pretrain Perch &  --- & \checkmark & --- & --- \\
        probe Perch &  --- & \checkmark & --- & \checkmark \\
        \bottomrule
      \end{tabular}
    \end{small}
  \end{center}
\end{table}

\begin{table}[t]
  \caption{Benchmark results on held out test data.}
  \label{tab:model-comparison}
  \begin{center}
    \begin{small}
      \begin{tabular}{lrrr}
        \toprule
        Model & Macro F1 & Macro AUC & Macro AP \\
        \midrule
        PULSE & 0.212 & 0.742 & 0.322 \\
        active PULSE & \textbf{0.338} & \textbf{0.838} & \textbf{0.556} \\
        pretrain Perch & 0.071 & 0.454 & 0.189 \\
        probe Perch & 0.329 & 0.832 & 0.439 \\
        \bottomrule
      \end{tabular}
    \end{small}
  \end{center}
\end{table}

\begin{figure}[tb]
  \vskip 0.2in
  \begin{center}
    \centerline{\includegraphics[width=\columnwidth]{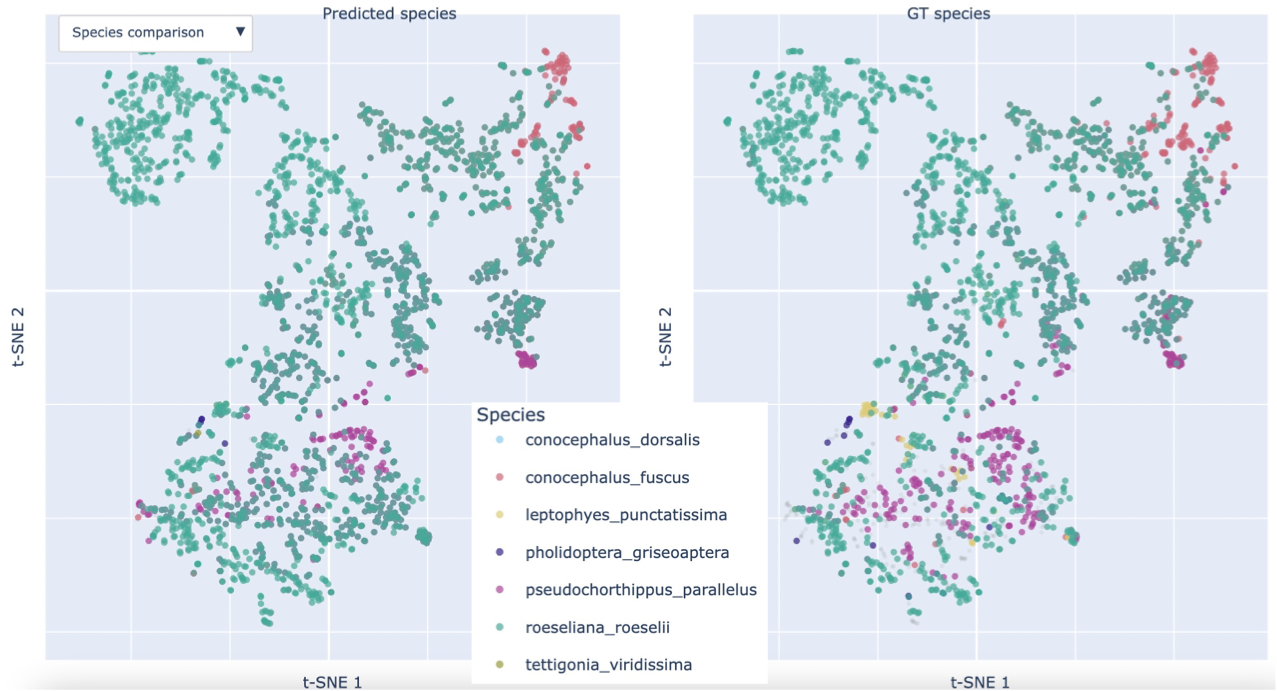}}
    \caption{
      Embedding visualisation tool for ecological analysis.
    }
    \label{fig:t_sne}
  \end{center}
\end{figure}

\begin{figure}[tb]
  \vskip 0.2in
  \begin{center}
    \centerline{\includegraphics[width=\columnwidth]{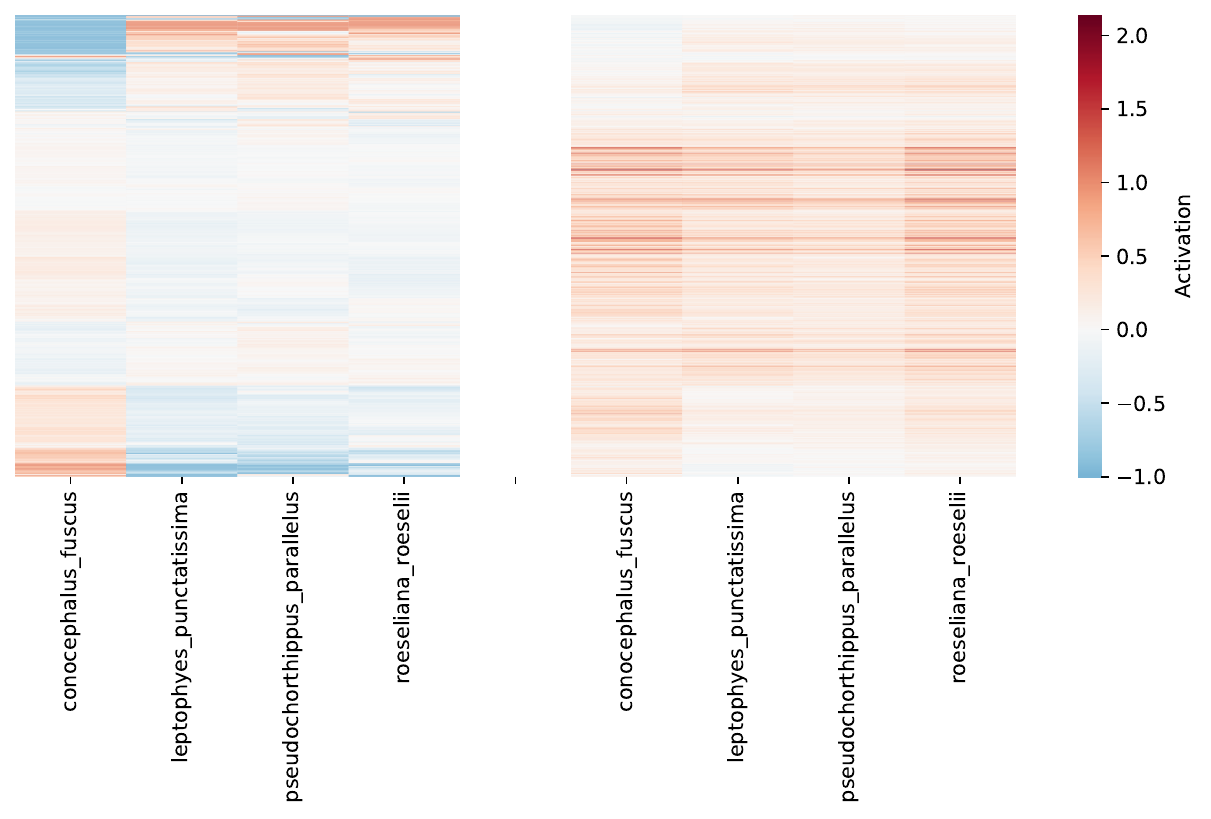}}
    \caption{
      Selectivity matrix  of 4 dominant species (left 4 columns) and examples of embeddings with individual species present (right 4 columns).
    }
    \label{fig:selectivity}
  \end{center}
\end{figure}

\begin{figure*}[tb]
    \centering

    \begin{subfigure}{0.24\textwidth}
        \centering
        \includegraphics[width=\linewidth]{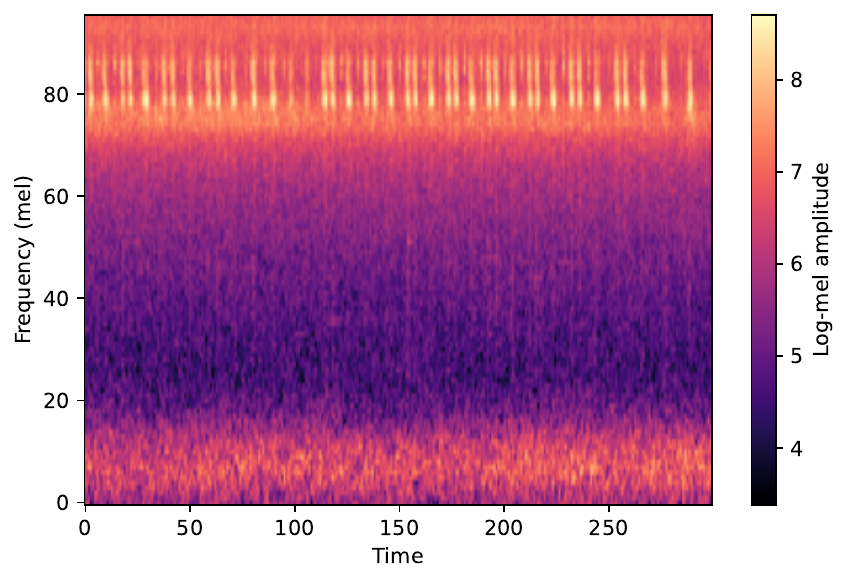}
        \caption{Mel spectrogram of \textit{Conocephalus fuscus} song.}
        \label{fig:spec_cf}
    \end{subfigure}
    \hfill
    \begin{subfigure}{0.24\textwidth}
        \centering
        \includegraphics[width=\linewidth]{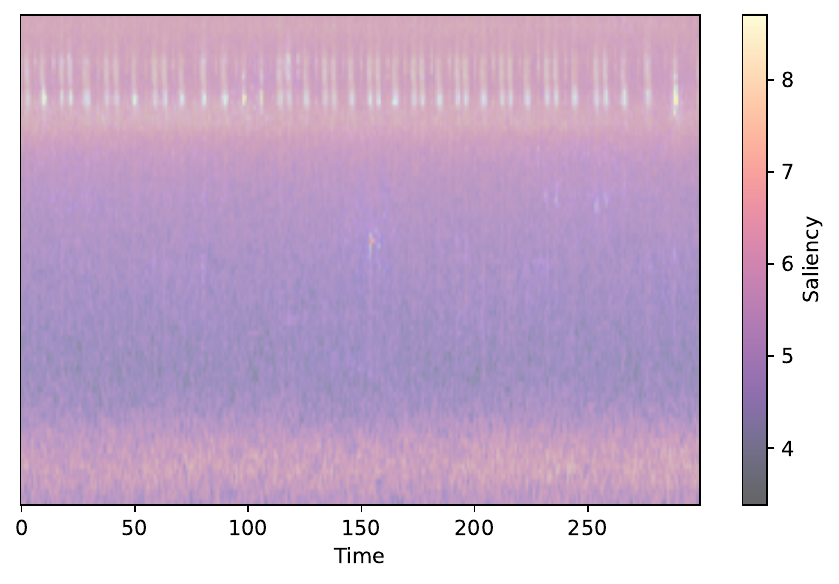}
        \caption{Saliency map of \textit{Conocephalus fuscus} song.}
        \label{fig:sal_cf}
    \end{subfigure}
    \hfill
    \begin{subfigure}{0.24\textwidth}
        \centering
        \includegraphics[width=\linewidth]{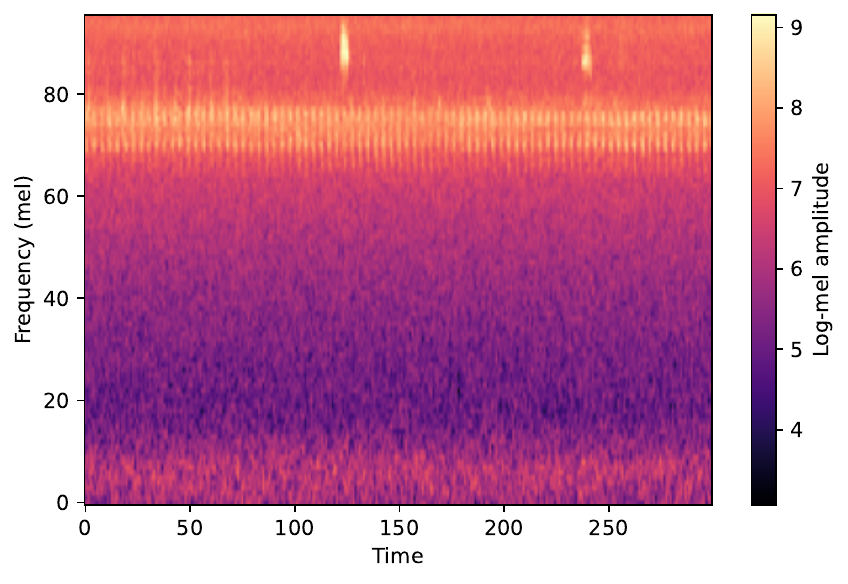}
        \caption{Mel spectrogram of \textit{Leptophyes punctatissima} song.}
        \label{fig:spec_lp}
    \end{subfigure}
    \hfill
    \begin{subfigure}{0.24\textwidth}
        \centering
        \includegraphics[width=\linewidth]{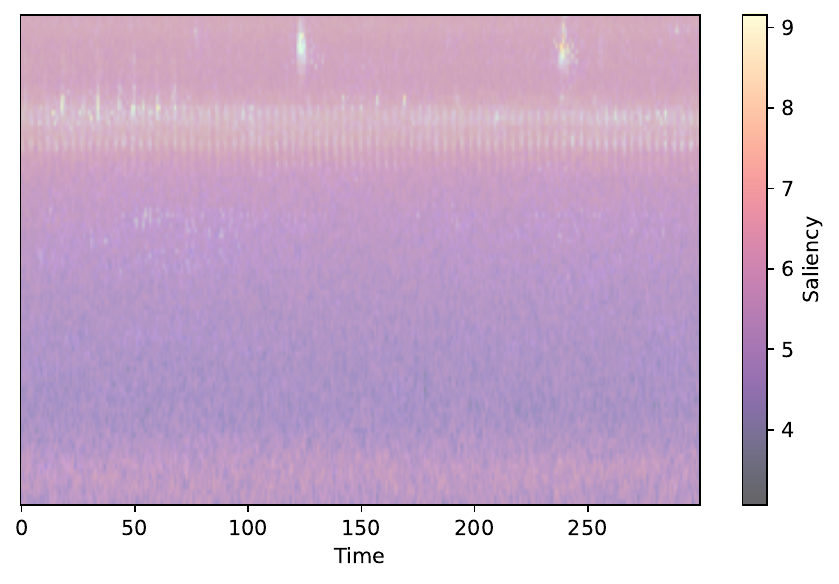}
        \caption{Saliency map of \textit{Leptophyes punctatissima} song.}
        \label{fig:sal_lp}
    \end{subfigure}

    \vspace{0.3cm}

    \begin{subfigure}{0.24\textwidth}
        \centering
        \includegraphics[width=\linewidth]{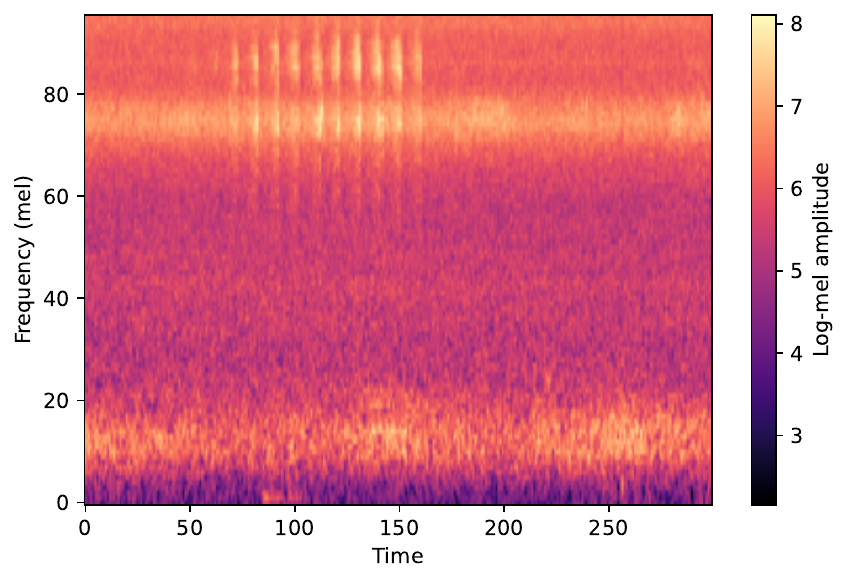}
        \caption{Mel spectrogram of \textit{Pseudochorthippus parallelus} song.}
        \label{fig:spec_pp}
    \end{subfigure}
    \hfill
    \begin{subfigure}{0.24\textwidth}
        \centering
        \includegraphics[width=\linewidth]{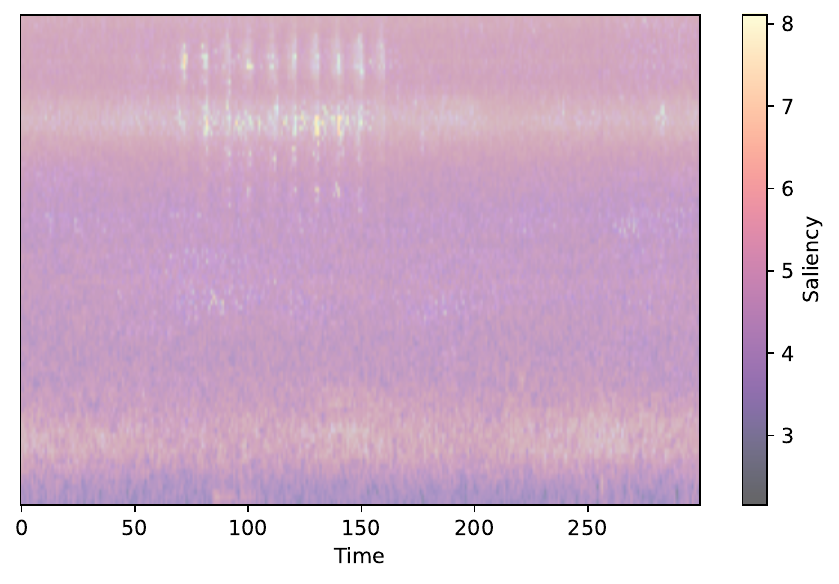}
        \caption{Saliency map of \textit{Pseudochorthippus parallelus} song.}
        \label{fig:sal_pp}
    \end{subfigure}
    \hfill
    \begin{subfigure}{0.24\textwidth}
        \centering
        \includegraphics[width=\linewidth]{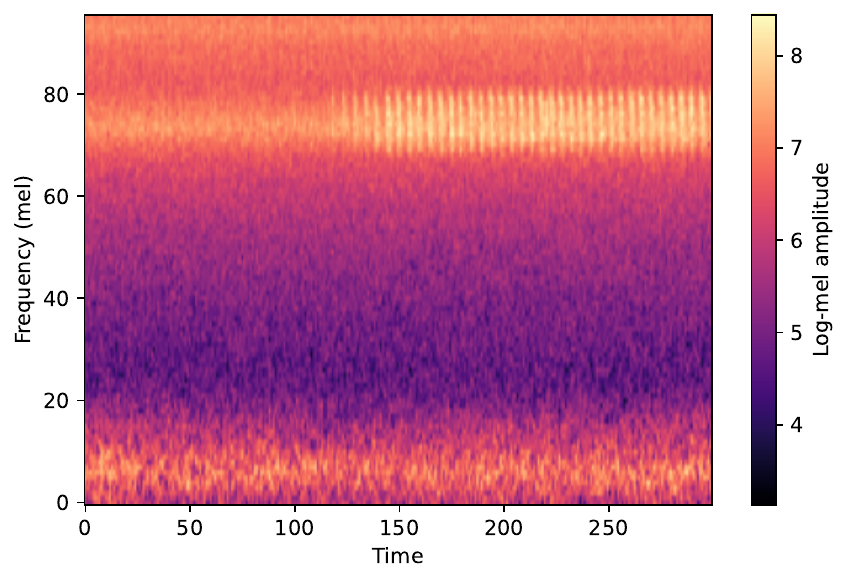}
        \caption{Mel spectrogram of \textit{Roeseliana roeselii} song.}
        \label{fig:spec_rr}
    \end{subfigure}
    \hfill
    \begin{subfigure}{0.24\textwidth}
        \centering
        \includegraphics[width=\linewidth]{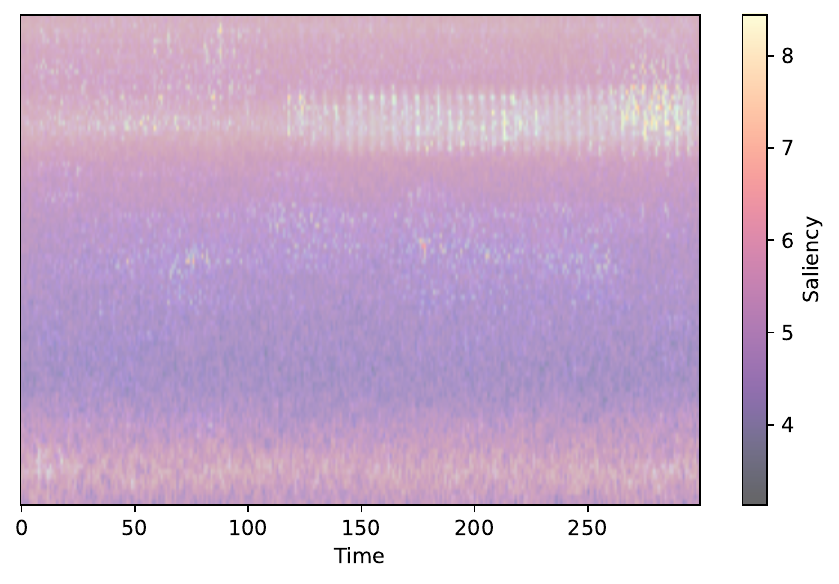}
        \caption{Saliency map of \textit{Roeseliana roeselii} song.}
        \label{fig:sal_rr}
    \end{subfigure}

    \caption{Saliency maps of single species recordings picturing all 4 dominant classes.}
    \label{fig:saliency_maps}
\end{figure*}

\begin{figure*}[tb]
    \centering

    \begin{subfigure}{0.32\textwidth}
        \centering
        \includegraphics[width=\linewidth]{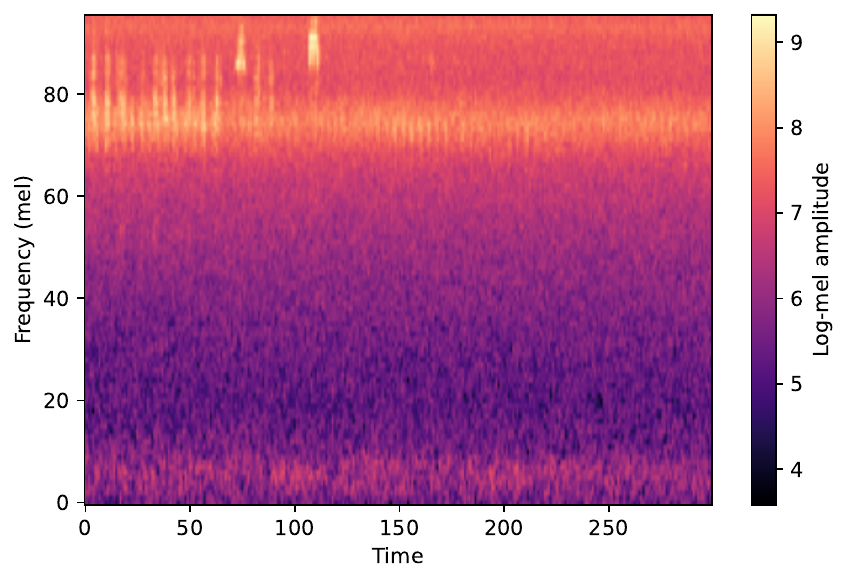}
        \caption{Mel spectrogram of mixed song recording.}
        \label{fig:spec_poly}
    \end{subfigure}
    \hfill
    \begin{subfigure}{0.32\textwidth}
        \centering
        \includegraphics[width=\linewidth]{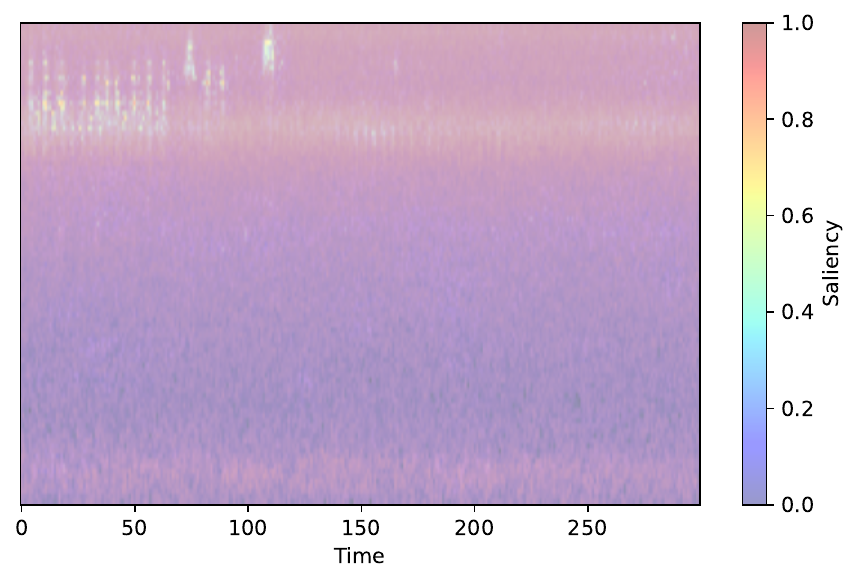}
        \caption{Saliency map of \textit{Leptophyes punctatissima} presence.}
        \label{fig:sal_poly_lp}
    \end{subfigure}
    \hfill
    \begin{subfigure}{0.32\textwidth}
        \centering
        \includegraphics[width=\linewidth]{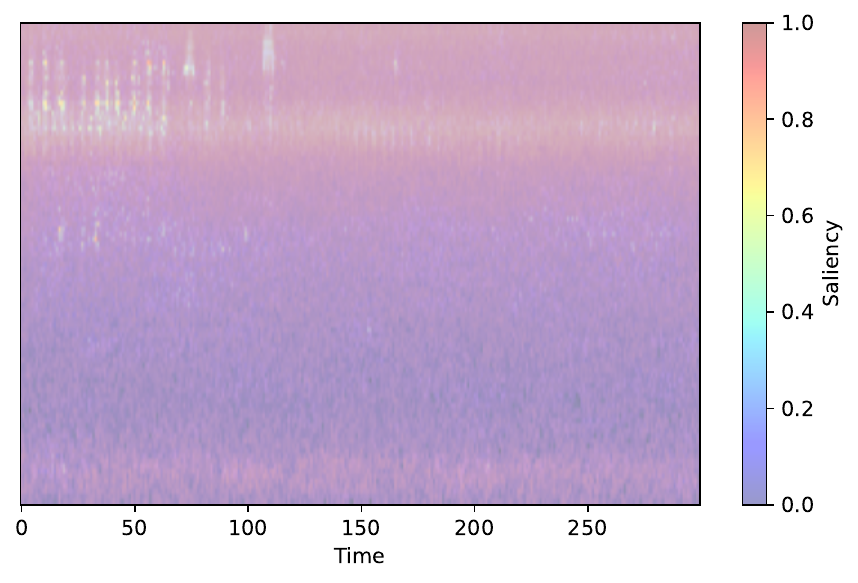}
        \caption{Saliency map of \textit{Pseudochorthippus parallelus} presence.}
        \label{fig:sal_poly_pp}
    \end{subfigure}

    \caption{Saliency maps of detected components of a mixed song recording. NNLS method on mean profiles detemines correct combination of species for this recording: \textit{Leptophyes punctatissima} and \textit{Pseudochorthippus parallelus}.}
    \label{fig:source_separation}
\end{figure*}

In this section, we evaluate PULSE on both classification performance and embedding quality; implementation details are in Appendix~\ref{app:implementation_details}. Test data was from 3 held-out sites entirely excluded from training -- including unlabelled data -- providing a direct assessment of model transferability to unseen environments.


\subsection{Benchmark Against Baselines}
As a baseline, we used Perch 2.0~\cite{van2025perch}, a recent extension of Perch~\cite{hamer2023birb}. Perch 2.0 has been shown to be very adaptable to new scenarios and species, even outperforming the specialist model for particular species~\cite{van2025perch}. 

We compared two versions of PULSE with two versions of Perch depending on whether they used the labelled field data or not (Table~\ref{tab:model-data}): \textit{PULSE} and \textit{pretrain Perch} only used labels from species library data, while \textit{active PULSE} and \textit{probe Perch} also used labels from the field data. Table~\ref{tab:model-comparison} shows the performance results. We found that PULSE significantly outperformed pretrain Perch, and active PULSE outperformed probe Perch. In both cases --- with and without local labels --- the specialist PULSE was more adaptive to the new environment than the general-purpose Perch.

PULSE generalises beyond Orthoptera: Appendix~\ref{app:more_results} presents results of PULSE extended to background classes such as cars and birds.

\subsection{Embedding Analysis}
\label{sec:emb_analysis}
As well as classification, the aim of our work was to build a model producing informative embeddings that will help ecologists with scientific discovery. We created an interactive visualisation of the embedding space (Figure~\ref{fig:t_sne}), which can colour data points by predicted species; and, if ground truth data is available, by species labels; the sites where the recording were made and time of day. This tool will help ecologists to explore and understand their data. Qualitative analysis of the embedding space with this visualisation tool is in Appendix~\ref{app:emb_analysis}.

To further characterise the embedding structure, we investigated species-specific information encoded in the embedding dimensions. For each of the four dominant species, we computed a selectivity index per embedding dimension — measuring how strongly each dimension responds to that species relative to all others~\cite{morcos2018importance} — and visualised the resulting selectivity matrix as a clustered heatmap (Figure~\ref{fig:selectivity}). Individual sample embeddings (on the right of Figure~\ref{fig:selectivity}) overlay naturally onto this structure, confirming that per-recording activations reflect their ground-truth species labels. One-way ANOVA across species confirmed that the learned embeddings were broadly informative: 891 of 1024 dimensions showed significant between-species variation ($F > 10$), with a median F-statistic of 39, indicating that discriminative structure was distributed across the full embedding space rather than concentrated in a small subset of dimensions.

We validated the selectivity profiles by computing Integrated Gradients saliency maps~\cite{sundararajan2017axiomatic} for single-species recordings, weighted by each species' selectivity profile (Figure~\ref{fig:saliency_maps}). The high-saliency regions corresponded closely to the key features of the call, namely the core frequency band and the areas of greatest amplitude, as seen in Figure~\ref{fig:sal_cf}, suggesting the model was well attuned to the distinguishing parts of the song.

We further applied this analysis to unmix recordings with multiple species (a proxy to source separation). For a recording with multiple species, we first decomposed its embedding via non-negative least squares (NNLS) onto per-species mean prototype vectors to identify which species were present. For each detected species, we then computed a selectivity-weighted saliency map over the input spectrogram. Figure~\ref{fig:source_separation} shows an example of two-species recording where NNLS correctly identifies the species. Figure~\ref{fig:sal_poly_lp} shows elevated saliency for both species, whereas Figure~\ref{fig:sal_poly_pp} correctly assigns low saliency to the other species' song. This asymmetry may reflect differences in how distinctly each species' call is represented, or patterns in the training data, such as common combinations with other species.

\section{Conclusion}

Using PULSE and active learning, we combined labelled and unlabelled data from newly available insect sound libraries to build a bioacoustic classifier tailored to UK Orthoptera. The learned embeddings revealed intraspecific behavioural differences and their environmental drivers, and enabled unmixing of overlapping calls using the embedding space alone. Ultimately, PULSE will facilitate large-scale insect PAM, providing vital information for conservation and habitat management.

\section*{Acknowledgements}
The authors are members of the Leverhulme Centre for Nature Recovery and are grateful for the support of the Leverhulme Trust (grant number RC-2021–076). We would also like to thank the anonymous reviewers for their constructive comments and suggestions, which helped improve this paper.

\section*{Impact Statement}

This paper presents work for which the goal is to advance the field of Machine Learning and the automatization of bioacoustic signals for monitoring ecological health. There are many potential societal consequences of our work, none which we feel must be specifically highlighted here.


\bibliography{literature}

@article{bennett2025recent,
  title={Recent technological developments allow for passive acoustic monitoring of Orthoptera (grasshoppers and crickets) in research and conservation across a broad range of temporal and spatial scales},
  author={Bennett, David and Nissen, Henning and Maschke, Marc Andre and Reck, Heinrich and Diek{\"o}tter, Tim},
  journal={Basic and Applied Ecology},
  volume={84},
  pages={147--157},
  year={2025},
  publisher={Elsevier}
}

@article{caruana1997multitask,
  title={Multitask learning},
  author={Caruana, Rich},
  journal={Machine learning},
  volume={28},
  number={1},
  pages={41--75},
  year={1997},
  publisher={Springer}
}

@article{faiss2023adaptive,
  title={Adaptive representations of sound for automatic insect recognition},
  author={Fai{\ss}, Marius and Stowell, Dan},
  journal={PLOS Computational Biology},
  volume={19},
  number={10},
  pages={e1011541},
  year={2023},
  publisher={Public Library of Science San Francisco, CA USA}
}

@article{fonseca2021fsd50k,
  title={{FSD50K}: an open dataset of human-labeled sound events},
  author={Fonseca, Eduardo and Favory, Xavier and Pons, Jordi and Font, Frederic and Serra, Xavier},
  journal={IEEE/ACM Transactions on Audio, Speech, and Language Processing},
  volume={30},
  pages={829--852},
  year={2021},
  publisher={IEEE}
}

@article{frommolt1996archive,
  title={The archive of animal sounds at the {H}umboldt-{U}niversity of {B}erlin},
  author={Frommolt, Karl-Heinz},
  journal={Bioacoustics},
  volume={6},
  number={4},
  pages={293--296},
  year={1996},
  publisher={Taylor \& Francis}
}

@article{funosas2026finely,
  title={A finely annotated dataset for the automated acoustic identification of European Orthoptera and Cicadidae},
  author={Funosas, David and Massol, Elodie and Bas, Yves and Schmidt, Svenja and Bennett, David and Arend, Dominik and Gebhard, Alexander and Barbaro, Luc and K{\"o}nig, Sebastian and Font, Rafael Carbonell and others},
  journal={Scientific Data},
  year={2026},
  publisher={Nature Publishing Group UK London}
}

@article{gibb2019emerging,
  title={Emerging opportunities and challenges for passive acoustics in ecological assessment and monitoring},
  author={Gibb, Rory and Browning, Ella and Glover-Kapfer, Paul and Jones, Kate E},
  journal={Methods in Ecology and Evolution},
  volume={10},
  number={2},
  pages={169--185},
  year={2019},
  publisher={Wiley Online Library}
}

@article{grill2020bootstrap,
  title={Bootstrap your own latent-a new approach to self-supervised learning},
  author={Grill, Jean-Bastien and Strub, Florian and Altch{\'e}, Florent and Tallec, Corentin and Richemond, Pierre and Buchatskaya, Elena and Doersch, Carl and Avila Pires, Bernardo and Guo, Zhaohan and Gheshlaghi Azar, Mohammad and others},
  journal={Advances in Neural Information Processing Systems},
  volume={33},
  pages={21271--21284},
  year={2020}
}

@article{hamer2023birb,
  title={Birb: A generalization benchmark for information retrieval in bioacoustics},
  author={Hamer, Jenny and Triantafillou, Eleni and Van Merri{\"e}nboer, Bart and Kahl, Stefan and Klinck, Holger and Denton, Tom and Dumoulin, Vincent},
  journal={arXiv preprint arXiv:2312.07439},
  year={2023}
}

@inproceedings{hershey2017cnn,
  title={{CNN} architectures for large-scale audio classification},
  author={Hershey, Shawn and Chaudhuri, Sourish and Ellis, Daniel PW and Gemmeke, Jort F and Jansen, Aren and Moore, R Channing and Plakal, Manoj and Platt, Devin and Saurous, Rif A and Seybold, Bryan and others},
  booktitle={2017 IEEE International Conference on Acoustics, Speech and Signal Processing (ICASSP)},
  pages={131--135},
  year={2017},
  organization={IEEE}
}

@article{kahl2021birdnet,
    author = {Kahl, Stefan and Wood, Connor M and Eibl, Maximilian and Klinck, Holger},
    title = {BirdNET: A deep learning solution for avian diversity monitoring},
    journal = {Ecological Informatics},
    year = {2021}
}

@article{balvanera2025whombat,
  title={Whombat: An open-source audio annotation tool for machine learning assisted bioacoustics},
  author={Balvanera, Santiago Mart{\'\i}nez and Mac Aodha, Oisin and Weldy, Matthew J and Pringle, Holly and Browning, Ella and Jones, Kate E},
  journal={Methods in Ecology and Evolution},
  volume={16},
  number={1},
  pages={19--28},
  year={2025},
  publisher={British Ecological Society}
}

@article{aodha2022towards,
  title={Towards a general approach for bat echolocation detection and classification},
  author={Aodha, Oisin Mac and Balvanera, Santiago Mart{\'\i}nez and Damstra, Elise and Cooke, Martyn and Eichinski, Philip and Browning, Ella and Barataud, Michel and Boughey, Katherine and Coles, Roger and Giacomini, Giada and others},
  journal={bioRxiv},
  pages={2022--12},
  year={2022},
  publisher={Cold Spring Harbor Laboratory}
}

@inproceedings{morcos2018importance,
  title={On the importance of single directions for generalization},
  author={Morcos, Ari S and Barrett, David GT and Rabinowitz, Neil C and Botvinick, Matthew},
  booktitle={International Conference on Learning Representations},
  year={2018}
}

@article{newson2017potential,
  title={Potential for coupling the monitoring of bush-crickets with established large-scale acoustic monitoring of bats},
  author={Newson, Stuart E and Bas, Yves and Murray, Ash and Gillings, Simon},
  journal={Methods in Ecology and Evolution},
  volume={8},
  number={9},
  pages={1051--1062},
  year={2017},
  publisher={Wiley Online Library}
}

@inproceedings{nuggehalli2025improved,
  title={Improved Algorithm for Deep Active Learning under Imbalance via Optimal Separation},
  author={Nuggehalli, Shyam and Zhang, Jifan and Jain, Lalit K and Nowak, Robert D},
  booktitle={International Conference on Machine Learning},
  pages={46815--46836},
  year={2025},
  organization={PMLR}
}

@article{riede2025acoustic,
  title={Acoustic monitoring for tropical insect conservation},
  author={Riede, Klaus and Balakrishnan, Rohini},
  journal={Philosophical Transactions of the Royal Society B: Biological Sciences},
  volume={380},
  number={1928},
  year={2025},
  publisher={The Royal Society}
}

@article{ross2023passive,
  title={Passive acoustic monitoring provides a fresh perspective on fundamental ecological questions},
  author={Ross, Samuel RP-J and O'Connell, Darren P and Deichmann, Jessica L and Desjonqu{\`e}res, Camille and Gasc, Amandine and Phillips, Jennifer N and Sethi, Sarab S and Wood, Connor M and Burivalova, Zuzana},
  journal={Functional Ecology},
  volume={37},
  number={4},
  pages={959--975},
  year={2023},
  publisher={Wiley Online Library}
}

@inproceedings{sundararajan2017axiomatic,
  title={Axiomatic attribution for deep networks},
  author={Sundararajan, Mukund and Taly, Ankur and Yan, Qiqi},
  booktitle={International Conference on Machine Learning},
  pages={3319--3328},
  year={2017},
  organization={PMLR}
}

@article{van2025perch,
  title={Perch 2.0: The bittern lesson for bioacoustics},
  author={van Merri{\"e}nboer, Bart and Dumoulin, Vincent and Hamer, Jenny and Harrell, Lauren and Burns, Andrea and Denton, Tom},
  journal={arXiv preprint arXiv:2508.04665},
  year={2025}
}

@article{vellinga2024xeno,
  title={Xeno-canto -- Orthoptera sounds from around the world},
  author={Vellinga, W},
  journal={Xeno-canto Foundation for Nature Sounds. Occurrence dataset https://doi.org/10.15468/76yp6j},
  pages={06--06},
  year={2024}
}

@article{walker2003effects,
  title={The effects of temperature and age on calling song in a field cricket with a complex calling song, {T}eleogryllus oceanicus (Orthoptera: Gryllidae)},
  author={Walker, Sean E and Cade, William H},
  journal={Canadian Journal of Zoology},
  volume={81},
  number={8},
  pages={1414--1420},
  year={2003},
  publisher={NRC Research Press Ottawa, Canada}
}

@article{xiao2022amresnet,
  title={{AMResNet}: An automatic recognition model of bird sounds in real environment},
  author={Xiao, Hanguang and Liu, Daidai and Chen, Kai and Zhu, Mi},
  journal={Applied Acoustics},
  volume={201},
  pages={109121},
  year={2022},
  publisher={Elsevier}
}
\bibliographystyle{icml2026}

\newpage
\appendix
\onecolumn
\section{Implementation details}
\label{app:implementation_details}
In this section we present all technical details of data and approaches we used.

\subsection{Field data collection}
\label{app:field_data_details}

Our field data was collected in summer 2024 across 10 sites in Oxfordshire, UK. On each site we installed 2 Audiomoths, a low-cost acoustic logger, that recorded 15 sec segments every minute at specified times: 4:00-6:00, 10:00-12:00, 17:00-19:00, 21:00-22:00 --- for 4-6 days at 96kHz sampling rate. Note that there were some failures and not all segments were recorded with one Audiomoth failing completely.

\subsection{List of species of interest}
Table~\ref{tab:species_list} gives the full list of 19 Orthoptera species used in our study. Table~\ref{tab:background_classes_list} presents the list of additional 6 background classes we used for extended PULSE in Appendix~\ref{app:more_results}. 

\begin{table}[tb]
  \caption{List of target Orthoptera species}
  \label{tab:species_list}
  \begin{center}
    \begin{small}
      \begin{tabular}{ll}
        \toprule
        Latin name & Common name \\
        \midrule
        \textit{Acheta domesticus} & house cricket \\
        \textit{Chorthippus albomarginatus} & lesser marsh grasshopper\\
        \textit{Chorthippus brunneus} & common field grasshopper\\
        \textit{Conocephalus dorsalis} & short-winged conehead\\
        \textit{Conocephalus fuscus} & long-winged conehead \\
        \textit{Gryllus bimaculatus} & two-spotted cricket\\
        \textit{Leptophyes punctatissima} & speckled bush-cricket\\
        \textit{Meconema meridionale} & southern oak bush cricket\\
        \textit{Meconema thalassinum} & oak bush-cricket\\
        \textit{Metrioptera brachyptera} & bog bush cricket\\
        \textit{Myrmeleotettix maculatus} & mottled grasshopper\\
        \textit{Omocestus rufipes} & woodland grasshopper\\
        \textit{Omocestus viridulus} & common green grasshopper\\
        \textit{Pholidoptera griseoaptera} & dark bush-cricket\\
        \textit{Pseudochorthippus parallelus} & meadow grasshopper\\
        \textit{Roeseliana roeselii} & European bush-cricket\\
        \textit{Stenobothrus lineatus} & stripe-winged grasshopper\\
        \textit{Stethophyma grossum} & large marsh grasshopper\\
        \textit{Tettigonia viridissima} & great green bush-cricket\\
        \bottomrule
      \end{tabular}
    \end{small}
  \end{center}
\end{table}

\begin{table}[tb]
  \caption{List of target Orthoptera species}
  \label{tab:background_classes_list}
  \begin{center}
    \begin{small}
      \begin{tabular}{ll}
        \toprule
        Background category \\
        \midrule
        Anthropophony \\
        Aves \\
        Bat \\
        Geophony \\
        Passeriformes \\
        Strigiformes \\
        \bottomrule
      \end{tabular}
    \end{small}
  \end{center}
\end{table}

\subsection{Public data details}
\label{app:data}

For labelled species library data we used a subset of ECOSoundSet data~\cite{funosas2026finely}, subsampling only data containing our selected list of species~(Table~\ref{tab:species_list}). ECOSoundSet has a mix of strong (a bounding box for each sound event) and weak (indicator that a recording contains a species) labels. For this work, we only worked with weak labels, therefore, all strong labels were converted to weak labels. Recordings in ECOSoundSet were of varying length, we split them into 3-sec clips. 

In addition to ECOSoundSet we use other online sources with the download script provided by~\cite{funosas2026finely}. Online sources include Xeno-canto~\citep{vellinga2024xeno}, iNaturalist, observation.org, SNSB (Bavarian State Collections of Natural History) and MinIO. Here we again used a subset of the data with the selected species~(Table~\ref{tab:species_list}). This data include only weak labels. Recordings in these datasets are of varying length with a label available for the whole recording only. For each recording we cropped up to 5 clips of 3 sec around the peaks\footnote{We used \texttt{scipy.signal.find\_peaks\_cwt()} function in Python to extract peaks.} (similar to Perch~\cite{van2025perch} and BirdNET~\cite{kahl2021birdnet}).  

Perch 2.0~\cite{van2025perch}, that we used as a baseline, was originally trained also on Xeno-Canto and iNaturalist, as well as additionally on the Tierstimmenarchiv~\cite{frommolt1996archive} and FSD50K~\cite{fonseca2021fsd50k} datasets. 

\subsection{Training details}
We held out three of the 10 sites for the test dataset to evaluate transferability, while data from remaining seven sites was randomly split between train and validation datasets in the ratio $8 / 1$, respectively. A part of the train data was labelled with the active learning approach (Section~\ref{sec:active-learning}), obtaining 552 labelled 15-sec recordings. We used the label taxonomy from ECOSoundSet~\cite{funosas2026finely}. Parts of the validation and test datasets were labelled at random, resulting in 67 and 197 labelled 15-sec recordings, respectively. These 15-sec recordings were split into 3-sec clips with overlap of 1 sec to be fed into the classification head of PULSE. 

All recordings (including unlabeled) were split into 3-sec clips to be used in the ecological prior head for comparison with the BirdNET embeddings. Additionally, clips of three seconds around one peak of the full 15-sec recording were extracted for the BYOL head. 

ECOSoundSet and open online data were split at random into train, validation and test datasets in the ratio $8 / 1 / 1$, respectively. Data from ECOSoundSet and online sources was used by the classifier head. 

We trained PULSE for 20 epochs, Adam optimiser with learning rate $1e-5$, exponential decay scheduled every 100,000 steps with rate 0.96, early stopping with patience 5 evaluated on validation data loss. 

We compared PULSE with pretrained Perch~\cite{van2025perch}. We used two versions of Perch. For the first version (\textit{pretrain Perch}), we used the model's direct predictions for our target classes. Since Perch was trained on the same Xeno-canto and iNaturalist datasets, it has all our selected species~(Table~\ref{tab:species_list}). Secondly, for \textit{probe Perch} we used embeddings from pretrained Perch and trained a linear probe connecting embeddings to logits for 19 selected species with binary cross-entropy loss. Linear probe Perch was trained for 50 epochs, Adam optimiser with learning rate $1e-3$, early stopping with patience 5 evaluated on validation data loss. 

For both methods we also used the validation dataset to tune classification threshold for positive predictions per class.

\subsection{Evaluation metrics}

\begin{table}[tb]
  \caption{Species with significant presence in test data}
  \label{tab:test_presence}
  \begin{center}
    \begin{small}
      \begin{tabular}{lc}
        \toprule
        Species & Number of test 15-sec recordings where it is present \\
        \midrule
        \textit{Conocephalus fuscus} & 41 \\
        \textit{Leptophyes punctatissima} & 14\\
        \textit{Pseudochorthippus parallelus} & 46\\
        \textit{Roeseliana roeselii} & 105\\
        \bottomrule
      \end{tabular}
    \end{small}
  \end{center}
\end{table}

For main results in Table~\ref{tab:model-comparison}, we used macro metrics --- AUC, AP --- over all classes that have non-zero support in labelled test data. For macro F1 we also included classes with only false positives. 

For embedding analysis (Section~\ref{sec:emb_analysis}), we focussed on 4 classes that had significant presence in the test data (Table~\ref{tab:test_presence}). The selectivity matrix, F-statistics and mean prototypes were computed over a balanced train data, whereas all examples in Section~\ref{sec:emb_analysis} are from the test data. For balancing of the train data we subsampled it to the size of the smallest class using a rarest-first, purity-prioritised procedure. Classes were processed in ascending order of sample count, so minority classes had first pick of examples. Within each class, candidates were prioritised by the number of active labels they carry — single-label recordings are selected first, followed by two-label recordings, and so on — with random selection within each purity tier to avoid index bias. Samples already selected for a previously processed class counted toward the quota of subsequent classes, preventing double-counting of overlapping multilabel recordings. 

We computed a selectivity matrix as $S_{d,k} = \frac{\mu_{d,k} - \mu_{d,\neg k}}{|\mu_{d,k}| + |\mu_{d,\neg k}| + \varepsilon}$, where $d$ is a dimensionality of the embedding space, $k$ is a class, $\mu_{d, k}$ is the mean for dimensionality $d$ and class $k$ among all recordings, $\mu_{d,\neg k}$ is the mean for dimensionality $d$ for all classes other than $k$. sch

It is worth mentioning, whilst \citet{morcos2018importance} caution that class selectivity in individual units does not causally explain network performance, our use of the selectivity index was purely diagnostic — we employed it as an interpretive tool to characterise the structure of learned representations rather than to make claims about which dimensions are necessary or sufficient for classification.

For Integrated Gradients~\cite{sundararajan2017axiomatic} we used zero baseline and 50 training steps. Note that the target signal is the selectivity-weighted sum of embedding dimensions rather than a class logit.

Mean prototype vectors used for unmixing multi-species recordings are just the mean of embeddings containing a species in the balanced train data. 

Species presence in mixed recordings was determined via NNLS: species with coefficients below 0.01 were deemed absent.

\section{Data source comparison}

The recordings from our field study sites provided extensive audio data for analysis. However, the passive nature of this method and use of relatively low-cost devices (AudioMoths) resulted in some noticeable differences compared with publicly available audio databases such as Xeno-canto. 

The annotated 15-sec clips contained a single species in 51\% of cases, two-species in 38\% and more than two species in the remaining 11\%. Xeno-canto recordings in comparison tended to be more focused on a single species, and whilst multi-species recordings did occur, they appeared to be much less frequent.

Recordings on Xeno-canto were generally collected using higher-specification microphones and were selectively collected at locations and timings to maximise the clarity and isolation of individual species' songs. In contrast, our field study recordings captured numerous additional sound sources alongside the target Orthoptera, including birds, wind, rain, and traffic.
    

\section{Additional Results}
\label{app:more_results}
Here we present the results of PULSE when we extended classes to include both selected Orthoptera species (Table~\ref{tab:species_list}) and background classes (Table~\ref{tab:background_classes_list}). Both ECOSoundSet data and our labelled field data contain training examples for background classes. Results are given in Table~\ref{tab:background-model-test}. We present both metrics computed for Orthoptera classes only (metrics with ``-O'' postfix) and for all classes. The results show that performance on Orthoptera classes remained competitive in comparison to PULSE trained on Orthoptera classes only (see \textit{active PULSE} in Table~\ref{tab:model-comparison}) and that performance on all classes remained at a comparable level. This suggests that PULSE can generalise to broader soundscape analysis, not only Orthoptera classification. 

\begin{table}[t]
  \caption{PULSE with background classes results on held out test data.}
  \label{tab:background-model-test}
  \begin{center}
    \begin{small}
      \begin{tabular}{p{3cm}cccccc}
        \toprule
        Model & Macro F1-O & Macro AUC-O & MacroAP-O & Macro F1 & Macro AUC & Macro AP \\
        \midrule
        PULSE with background classes & 0.383 & 0.827 & 0.465 & 0.363 & 0.749 & 0.458 \\
        \bottomrule
      \end{tabular}
    \end{small}
  \end{center}
\end{table}

\section{Qualititative Embedding Analysis}
\label{app:emb_analysis}

This section provides analysis based on interactive embedding visualisation of the labelled train data.

\subsection{Embedding Space Reveals Clear Separation for Two of the Three Most Common Species but Wider Dispersion for the Third}

\begin{figure}[tb]
    \centering
    \includegraphics[width=1\linewidth]{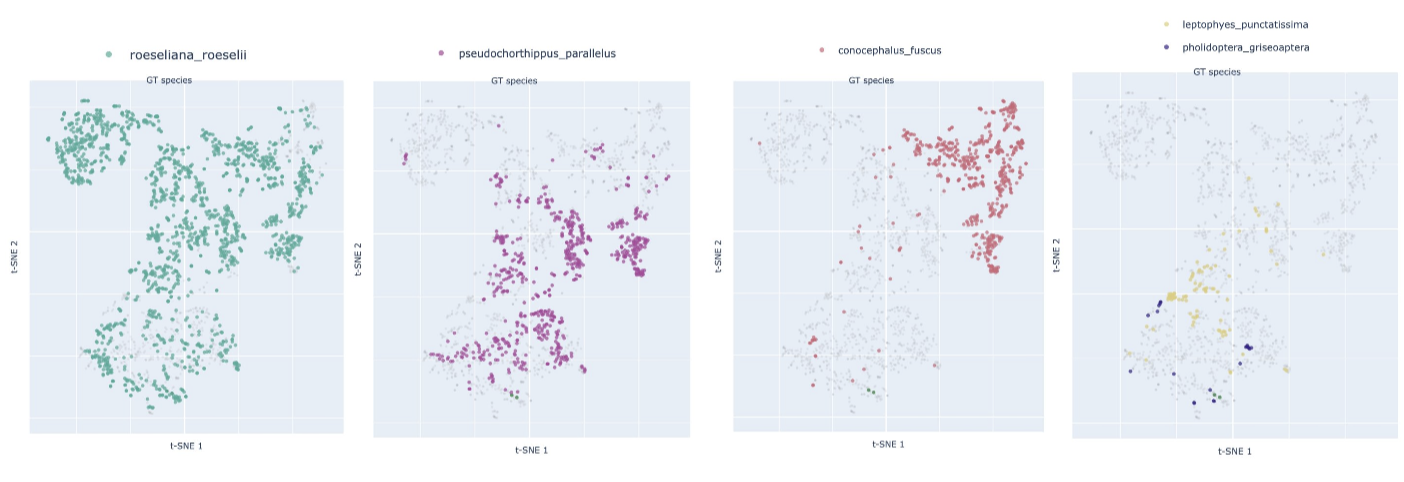}
    \caption{Embedding visualisation of labelled train data. The three left-hand plots are coloured by presence of the three most prevalent species; the right-most plot shows the less common species.}
    \label{fig:t-sne_classes}
\end{figure}

Three species were prevalent across the annotated dataset: \textit{Roeseliana roeselii} (Rr), \textit{Conocephalus fuscus} (Cf), and \textit{Pseudochorthippus parallelus} (Pp). A further two species were less frequently observed: \textit{Pholidoptera griseoaptera} (Pg) and \textit{Leptophyes punctatissima} (Lp)\footnote{\textit{Pholidoptera griseoaptera} has a very little support, only 1 recording in the test data, whereas \textit{Leptophyes punctatissima} has enough support to be included in analysis in Section~\ref{sec:emb_analysis}} (Figure~\ref{fig:t-sne_classes}). Among the three dominant species, Pp and Cf occupied relatively distinct regions of the embedding space, with some overlap where both species co-occurred within a single embedding (see Appendix~\ref{app:multi_clusters}). Embeddings containing Rr, the most frequent of the three, were distributed more widely. Some of the contributing factors for this are discussed in subsequent sections.

\subsection{Embeddings with Multi-Species Composition Showed a Tendency to Cluster}
\label{app:multi_clusters}

\begin{figure}[tb]
    \centering
    \includegraphics[width=0.5\linewidth]{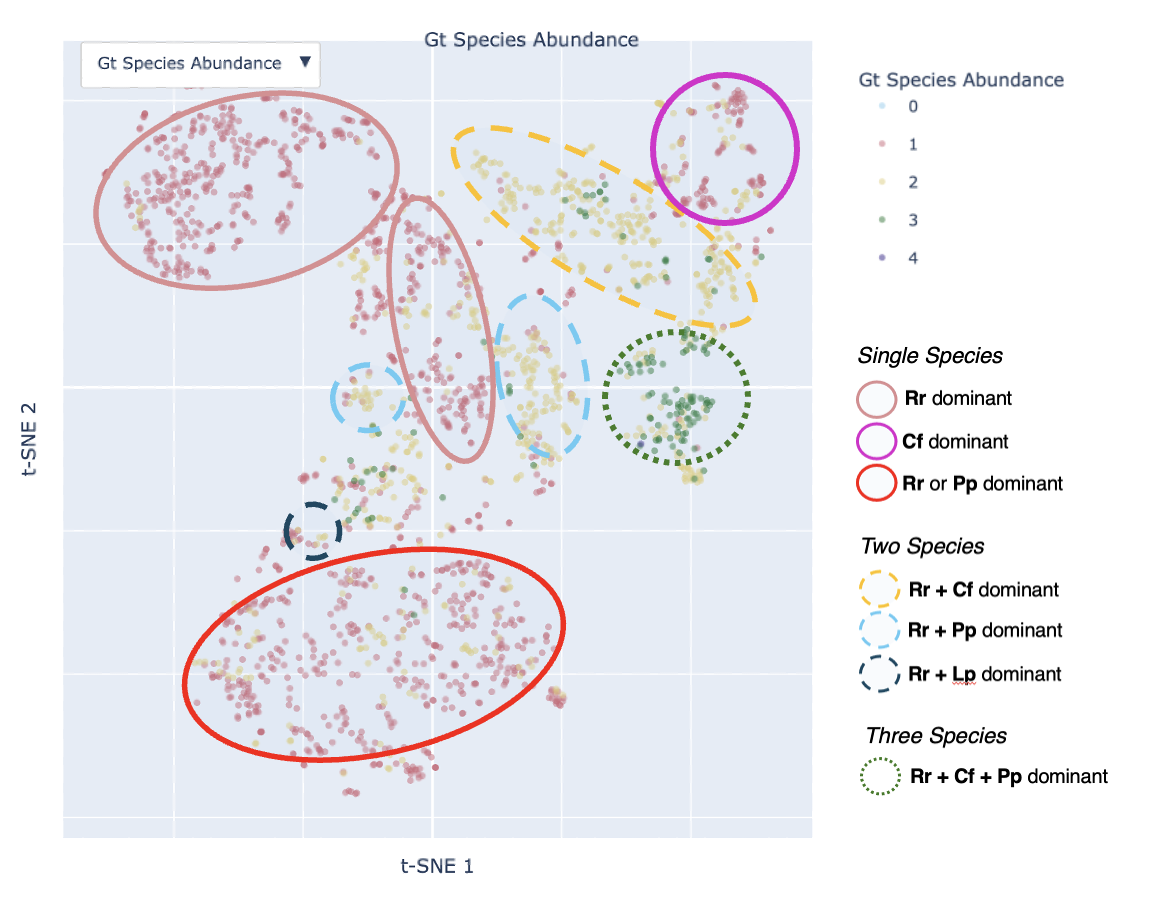}
    \caption{Embeddings coloured by the number of species present in each embedding. Manually drawn ellipses indicate visually identified clusters, with line style and colour denoting the predominant species combination within each cluster, as determined from the ground truth annotations.}
    \label{fig:t-sne_abundance}
\end{figure}

Clustering of embeddings appeared to reflect both the number of species present and their particular combinations. Furthermore, multi-species embeddings tended to be surrounded by single-species embeddings of one constituent species, suggesting that one species acted as the dominant feature in the embedding (Figure~\ref{fig:t-sne_abundance}). For instance, the cluster of Cf in the top-right of Figure~\ref{fig:t-sne_abundance} (solid purple ellipse) is next to a cluster of predominantly Cf and Rr immediately below (yellow dotted ellipse). 

\subsection{Multi-Species Embeddings with Varying Composition Can Cluster by Recording Site}

\begin{figure}[tb]
    \centering
    \includegraphics[width=0.75\linewidth]{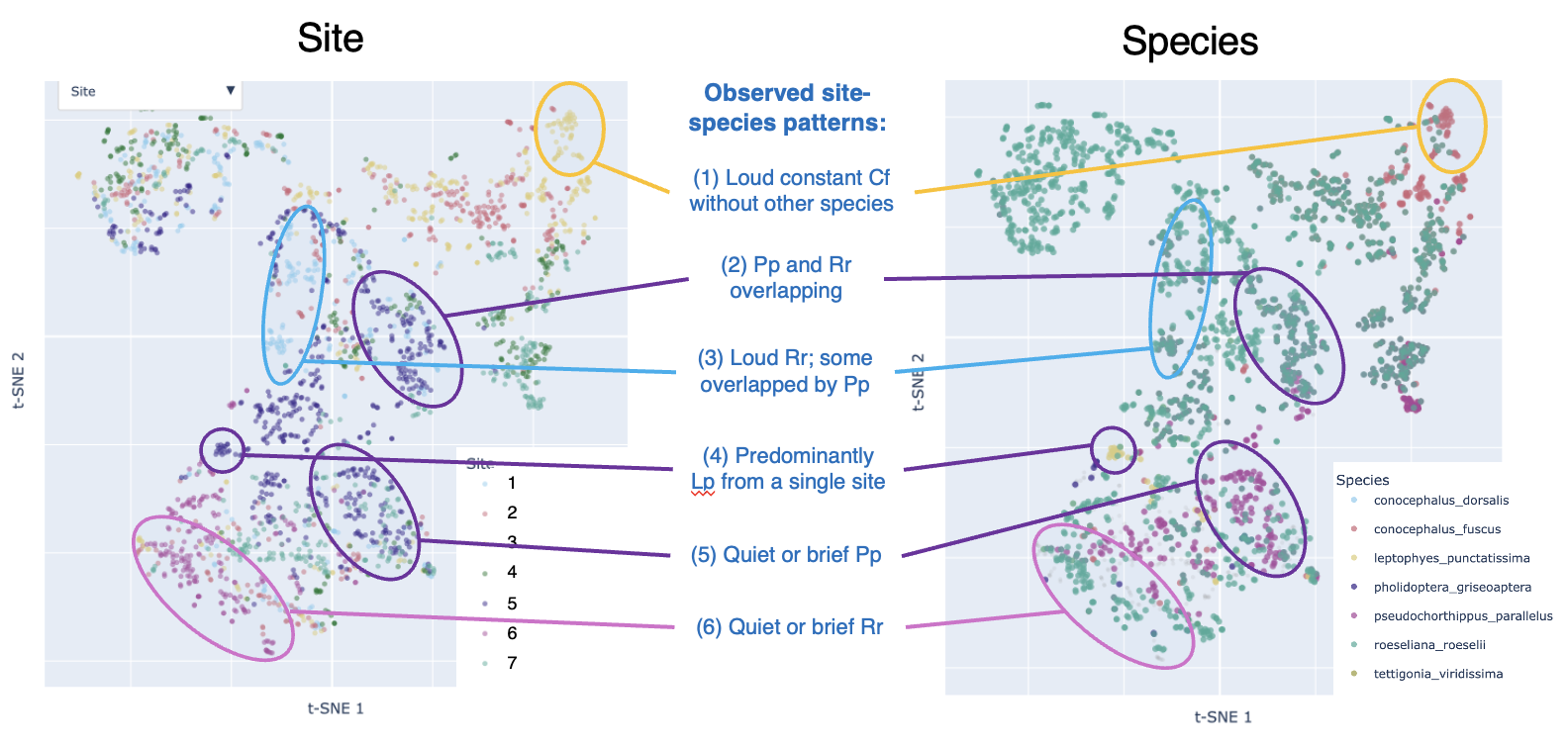}
    \caption{Embeddings coloured by recording site (left), with ellipses highlighting visible site clusters, linked to a plot coloured by species (right), illustrating how some site clusters correspond to particular species combinations or song characteristics.}
    \label{fig:t-sne_site}
\end{figure}

Clustering of different species in certain areas of the embedding space aligned in some instances with the recording site. For example, the top-right corner of the plot contains many single-species Cf embeddings, almost all from a single site. Analysis of the spectrograms indicated that these were predominantly loud Cf songs that persist throughout the clips (Figure~\ref{fig:t-sne_site}). Further examples of site-specific clusters, linked to particular species combinations or song characteristics (e.g. constant or brief), are illustrated in Figure~\ref{fig:t-sne_site}.

\subsection{Single-Species Embeddings, Such as \textit{Roeseliana Roeselii}, Also Cluster due to Variations in Song.}

\begin{figure}[tb]
    \centering
    \includegraphics[width=0.75\linewidth]{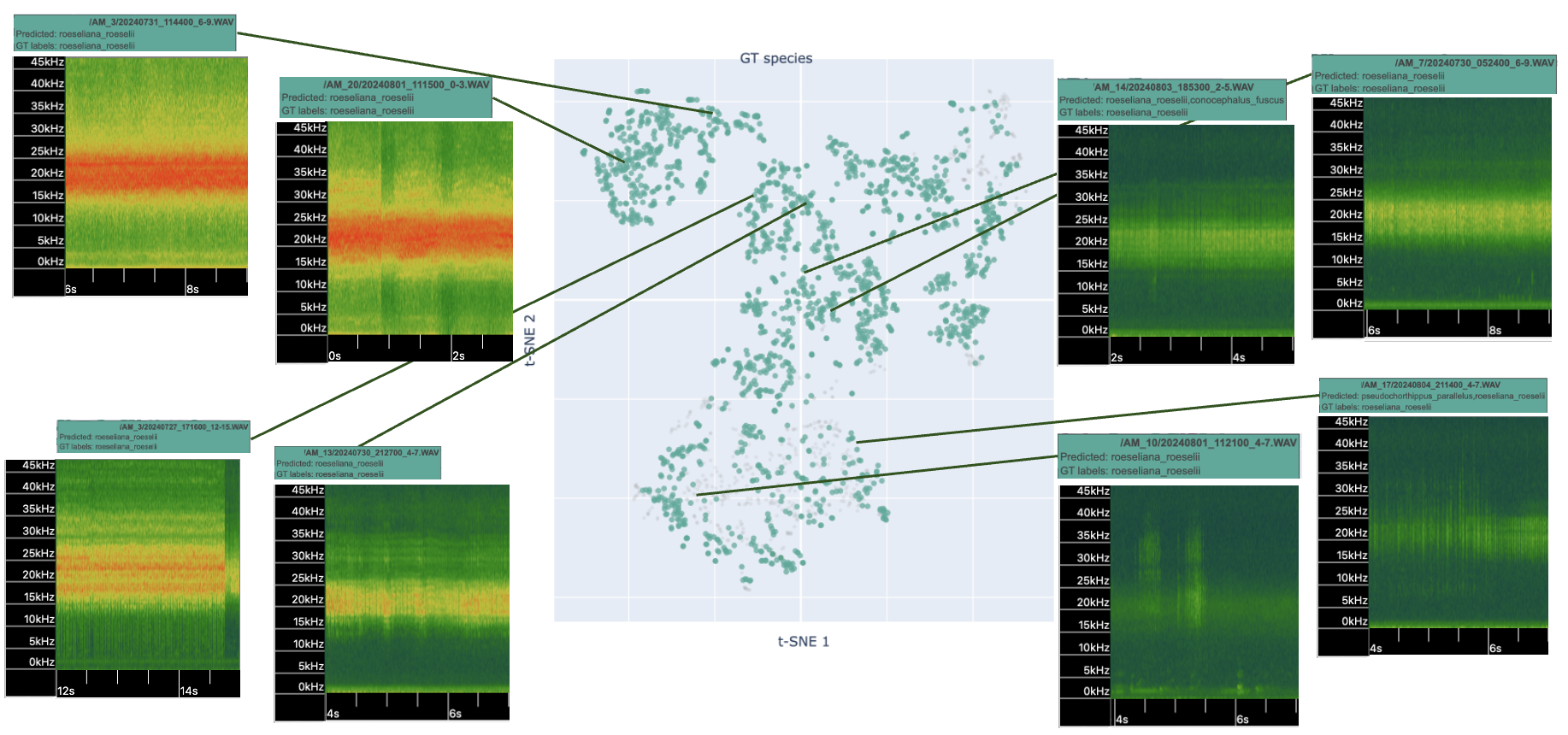}
    \caption{Roeseliana roeselii embeddings with a random selection of points linked to their spectrograms, illustrating the variation in song features across the embedding space.}
    \label{fig:t-sne_roeseliana}
\end{figure}

\textit{Roeseliana roeselii} clustered by multi-species combination (see Appendix~\ref{app:multi_clusters}),  but the spread of their embeddings across most of the plot appeared to be influenced by multiple environmental and ecological factors that vary by site and individual. The most noticeable of these were the loudness and the speed (or pulse rate) of the song, which both vary widely as reflected in the spread of embeddings towards the edges of the plot. The duration of the song within the clip and number of individuals singing appear to also have an effect. This was also observed for Pseudochorthippus parallelus, however, this was less clear for other species (Figure~\ref{fig:t-sne_roeseliana}). 

\subsection{Time of Day Produces Some Clear Clusters That Appear to Reflect Song Variation, Particularly in Chirp Rate, for Example \textit{Pseudochorthippus Parallelus}}

\begin{figure}[tb]
    \centering
    \includegraphics[width=0.75\linewidth]{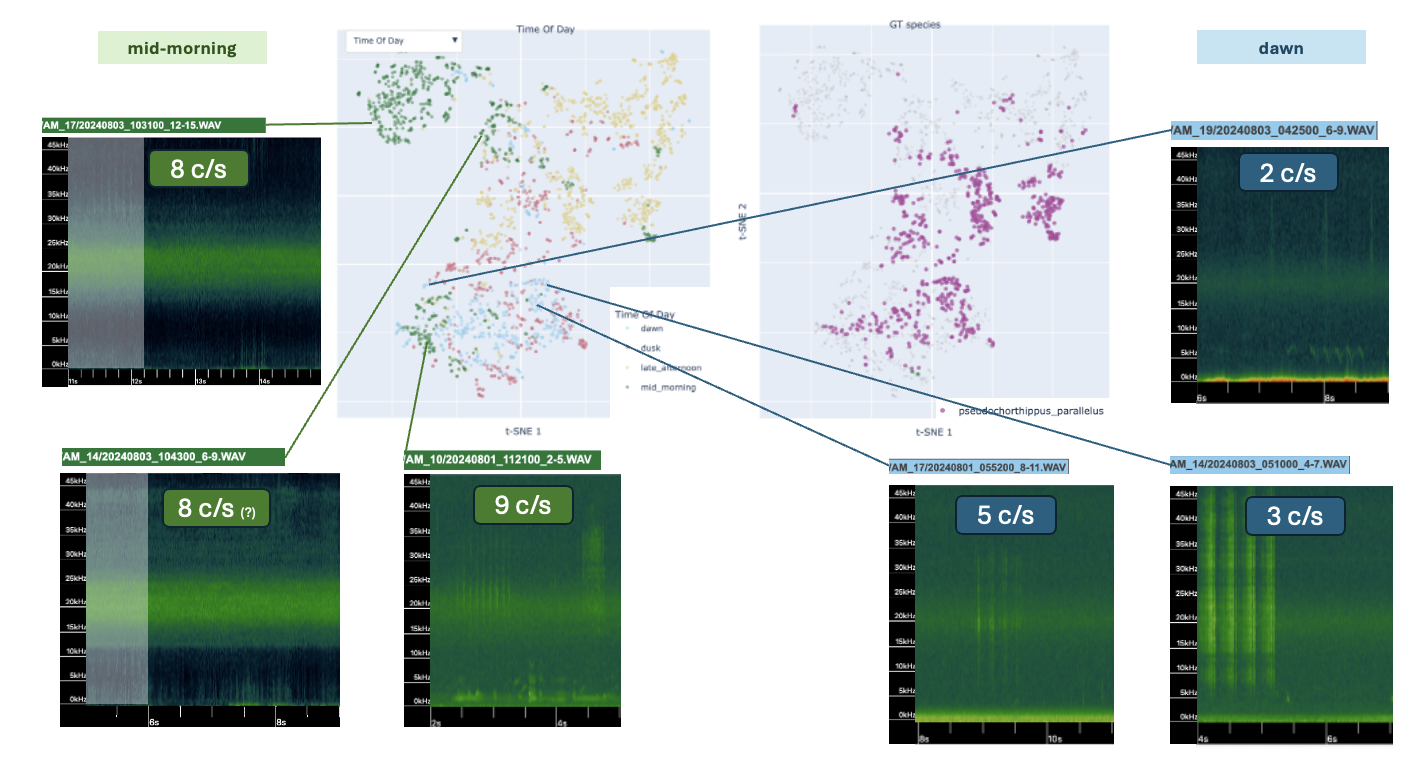}
    \caption{Embeddings with Time of Day for all annotations (left) and the spread of embeddings with \textit{Pseudochorthippus parallelus} (right). The selected spectrograms show three from the mid-morning and three from around dawn with the approximate chirp rate added to each plot. Note the left hand two plots show the preceding 1-second as the selected clip only catches the end of the song.}
    \label{fig:t-sne_time_of_day}
\end{figure}

Temperature has been identified as a key factor influencing the chirp rate of various Orthoptera species~\cite{walker2003effects}. This is largely attributable to temperature variation and was observed in our data for \textit{Pseudochorthippus parallelus} and \textit{Roeseliana roeselii} in particular. The dawn and dusk chirp rates of these species were noticeably slower than mid-morning and late-afternoon rates (Figure~\ref{fig:t-sne_time_of_day}). 

\subsection{AudioMoth Tonal Band (Around 15-25kHz) Can Result in Clustering Together of Brief or Quiet Songs From Different Species}

\begin{figure}[tb]
    \centering
    \includegraphics[width=0.75\linewidth]{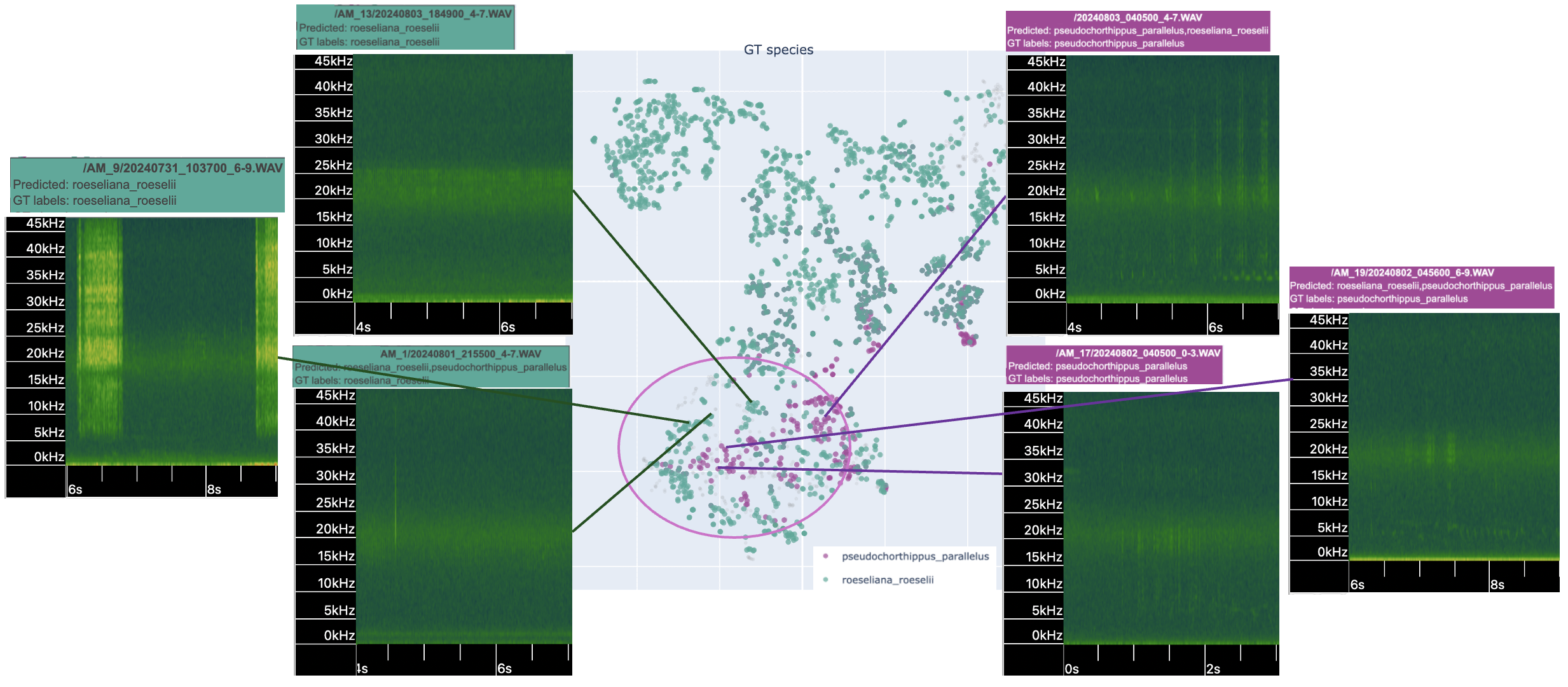}
    \caption{Single species embeddings with a focus on spectrograms where the embeddings with a single species of either Roeseliana roeselii (green) or Pseudochorthippus parallelus (purple) are placed in a similar area. The AudioMoth tonal band can be seen between 15-25 kHz in all the spectrograms.}
    \label{fig:t-sne_tonal}
\end{figure}

A consistent tonal band, visible as a brighter region on spectrograms between 15–25kHz, appeared throughout the field dataset. This is a known characteristic of AudioMoth recordings and was present regardless of environment or recording settings. As this frequency range overlaps with the core calling frequencies of some Orthoptera species, it could reduce the distinctiveness of their songs on the spectrogram. For distant or brief calls, the tonal band became the more prominent feature, and this appeared to be the common factor linking otherwise dissimilar songs in the same cluster. This is reflected in certain areas of the plot, where embeddings containing a single species, either Rr or Pp, sat in close proximity despite the two songs being acoustically distinct, suggesting that quieter calls were harder to differentiate at the embedding level (Figure~\ref{fig:t-sne_tonal}).


\end{document}